\documentclass[times, twoside]{zHenriquesLab-StyleBioRxiv}
\usepackage{blindtext}
\usepackage[overload]{textcase}
\usepackage{svg}

\usepackage{enumitem,amssymb,amsthm,amsmath,  mathdots, yhmath, cancel, color}

\newlist{todolist}{itemize}{2}
\setlist[todolist]{label=$\square$}
\usepackage{pifont}
\begin{document}

    \bibliographystyle{zHenriquesLab-StyleBib}
    
\leadauthor{Bocșe \& Jinga}


\title{The Immersion of Directed Multi-graphs in Embedding Fields. Generalizations.}
\shorttitle{The modeling of particle-field duality under assumption of perfect information minimality}

\author[1 2]{Bogdan Bocșe}
\author[1 3]{Ioan Radu Jinga}
\affil[1]{Knosis.AI, Bucharest, Romania}
\affil[2]{Envisage.AI, Bucharest, Romania}
\affil[3]{Jiratech, Bucharest, Romania}

\maketitle

\begin{abstract}

The purpose of this paper is to outline a generalized model for representing hybrids of relational-categorical, symbolic, perceptual-sensory and perceptual-latent data, so as to embody, in the same architectural data layer, representations for the \textit{input}, \textit{output} and \textit{latent} tensors. This variety of representation is currently used by various machine-learning models in computer vision, NLP/NLU, reinforcement learning which allows for direct application of cross-domain queries and functions. This is achieved by endowing a directed \textit{Tensor-Typed  Multi-Graph} with at least some edge attributes which represent the embeddings from various latent spaces, so as to define, construct and compute new similarity and distance relationships between and across tensorial forms, including visual, linguistic, auditory latent representations, thus stitching the logical-categorical view of the observed universe to the Bayesian/statistical view.

\end {abstract}


\begin{corrauthor}
    bogdan\at envisage.ai, radu.jinga@jiratech.com
\end{corrauthor}

\subsection*{Introduction}

The convergence of decentralized applications, deep learning and semi-autonomous machines is making it mandatory for the future enterprise of the 2020s to prepare their architectures for the incoming waves of business demand and opportunity. Specifically, the design priorities and hardware restriction which shaped the last 30 years of relational (i.e. often called by the misnomer "SQL") and non-relational database (i.e. often called by the misnomer "NoSQL") history bare little relevance for the data life cycle and access pattern required in the decade of Artificially Augmented Intelligence (AAI) and "in silico" direct and latent perception. 

For this purpose, we explore a set of novel concepts for data store and data warehouse architectures, starting from the low-level analytical, functional and canonical considerations of the many embodiment of data (i.e.\textit{entropy}) and data structures (i.e. relationships between various types of entropy). As the general formal model may seem counter-intuitive for conventional business use cases, we shall also provides a few examples and embodiment of the Tensor/Typed-Multi-graph in real-world machine learning scenarios. We shall thus outline a generalized model for representing hybrids of relational, symbolic, perceptual-sensory and perceptual-latent data, so as to embody, in the same architectural data layer, the symbolic and sensory input, output and latent space representations (i.e. embeddings spaces and/or fields) used by various machine-learning models in computer vision, in NLP/NLU, and in reinforcement learning. For this reason, we are going relevant work from disparate fields, including topology, graph and multi-graph theory, probability distributions and computation theory. This purpose-agnostic design allows for direct application of cross-domain queries and functions, \textit{without requiring human intervention in the application of change requests}. This is achieved by endowing a directed multi-graph with \textit{tensorial or typed} values at least for some edge attributes, which thus become delegate with representing the typed-embeddings from various latent spaces, so as to compute similarity and distance relationships between and across tensorial forms, including visual, linguistic and auditory latent space embeddings.

\subsection*{Definition of Entities}
 The generic multi-graph $G=G\{V; E\}$ is defined by the typed vertices
 
\begin{equation*}
V=\left \{ v | v_{id}=id_{v}(...) ; \textit{type(v)} \in T_{V} ; [K_{T_{v}} \in D_{T_{v}} \rightarrow O \in D_{K_{v}(T_{v})}  ] \right \}
\end{equation*}

and the typed edges
\begin{equation*}
E=\left \{ e | e_{id}=id_{e}(...) ; \textit{type(e)} \in T_{E} ; [K_{T_{e}} \in D_{T_{e}} \rightarrow O \in D_{K_{e}(T_{e})}  ] \right \}
\end{equation*}

 with the following assumptions of notation:
 
 \begin{itemize}
\item $V$ is the set of vertexes, with individual items $v$
\item $E$ is the set of edges, with individual items $e$
\item the function $type( ... )$ returns the type of individual edge or vertex

\item All $id_{e}(...)$ and $id_{v}(...)$ are embodiment-dependent, time-dependant, possibly-stateful, functions that generate pseudo-unique pseudo-random identifiers, with or without causally-dependant validation of absolute unicity. The idenfiers  bit arrays generated to be pseudo-unique across both sets V and E, generated by juxtaposing a random number to the nanosecond clock of the machine. To avoid any confusion and to only allow for strictly-type comparisons (i.e. not allowing implicit comparison between edge IDs and vertex IDs), a specific bit-sequence will be prefixed to vertex IDs ( $ID_{v}$ ) and a different bit-sequence to edge IDs ( $ID_{e}$ )

\item The attribute values associated to the keys $K_{T_{v}}$ and $K_{T_{e}}$ are represented as the observable $O$, which is restricted to its type-specific dictionary

 \item the set $K_{T_{v}}$ refers to all keys allowed for the vertex type $T_{v}$ which is \textbf{<defined by>} and \textbf{<restricted to>} the vertex-type-specific dictionary $D_{T_{v}}$, of possible values. 
 
 \item the set $K_{T_{e}}$ refers to all keys allowed for the edge type $T_{e}$ which is \textbf{<defined by>} and \textbf{<restricted to>} the edge-type-specific dictionary $D_{T_{e}}$, of possible values. 
 
 \item The dictionary-restriction $D_\{ ... \}$ for each key of each vertex/edge should be seen as \textbf{<a set of types>}, including enumerations, primitive types (fixed point representation, floating point representations), composite types (object hierarchies), \textit{array}, \textit{matrices} and \textit{tensor types} (including strings, images, animations and higher-dimensional tensors, across a variety of partially ordered and partitioned dictionaries quantization). Let us assume, for the sake of simplicity, that we only allow those types which have \textbf{at least one} metric function  \textit{well-defined} and \textit{computable in a tractable way}. Example of such metric functions include, without limitation:  a norm, a distance, a geodesic, a calculable path-finding cost function (including \textit{eg. the Dijkstra algorithm or the A-Star* heuristic}), a definition of difference or variation (finite difference), possibly warped across space, time, or frequency dimensions  (\textit{eg. Dynamic Time Warping - DTW \cite{salvador-chan}, Correlation Optimized Warping - COW \cite{tomasi}, Levenshtein edit-distance \cite{levenshtein} }).
 
 \end{itemize}
 
  The following example represents a multi-graph with several types of directed edges and several types of vertexes:

 \begin{figure}[!t]
    \centering
    \includegraphics[width=\linewidth]{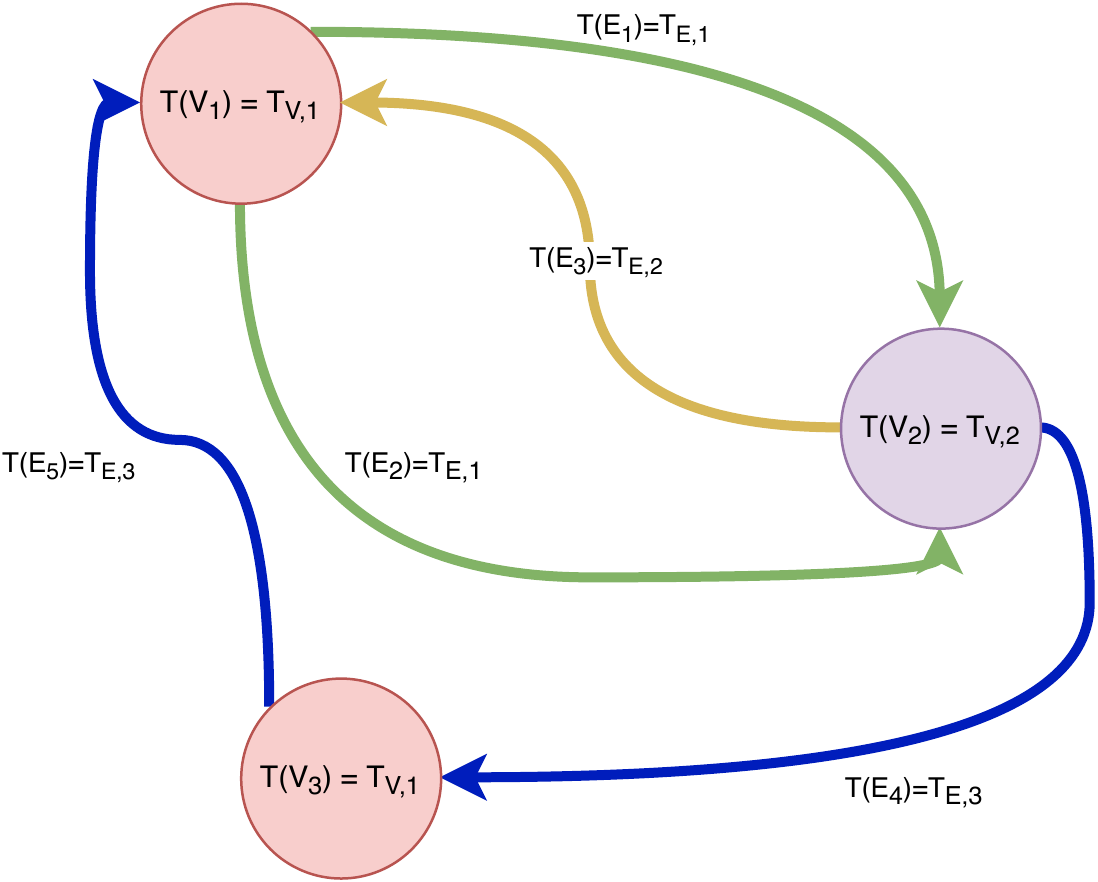}
    \caption{\textbf{Several directed edges ($E_1$, $E_2$, $E_3$, $E_4$ and $E_5$) of different types ($T_{E,1}$, $T_{E,2}$ and $T_{E,3}$) spanning between vertexes ($V_1$, $V_2$ and $V_3$), which themselves have different types ($T_{V,1}$ and $T_{V,2}$) }}
    \label{fig:typesOfEdgesAndVerties}
 \end{figure}
 
 The purpose of this \textit{tensor-typed multi-graph} is to dismount limiting assumptions on the mostly algebraic and logic functions computable by conventional-relational data stores. Concretely, by means referenced and described in this paper, the consumers (i.e. user or services) of such a multi-modal tensor-typed graph will be able not only to operate and compute with existing node types and edge types, but will also be able to define new types (classes) of objects and new types of (relationships) between objects, in the pursuit of some meta-objective functions, heuristics or estimators (predictability, safety, performance, robustness), in good awareness of cost and capacity constraints, and with offering the opportunity for human non-expert users to either \textbf{vote}, \textbf{bet} or \textbf{bid} on the changes of the quality-cost trade-off.

 \subsection*{Representing the probability-space-time-energy (pSTEM) tensor space}
 
\begin{figure}
    \makebox[\linewidth]{
        \includegraphics[width=\linewidth]{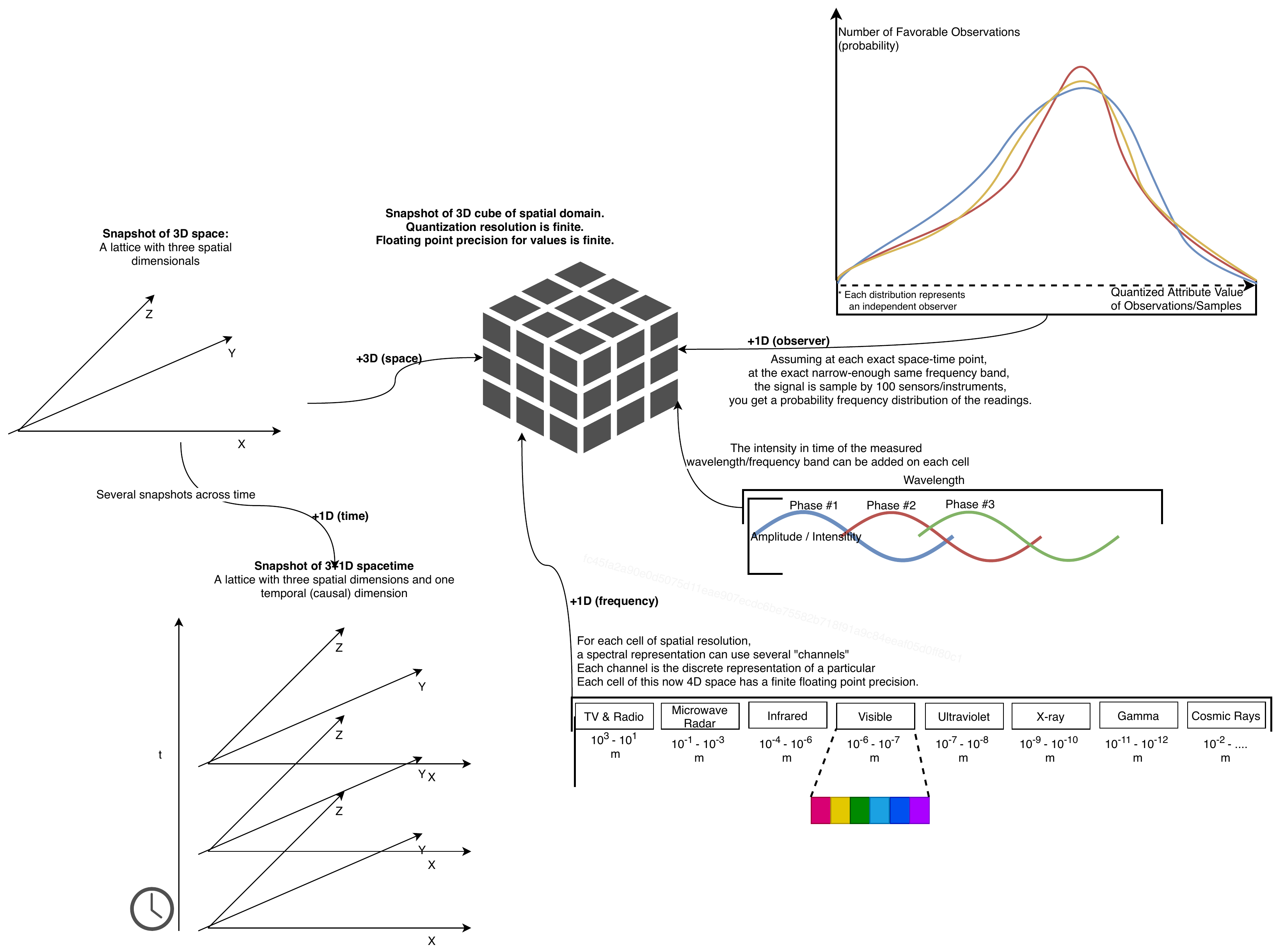}
    }
       \caption{\textbf{Representation of a tensorial volume} A generalized visualization of the input field, with three spatial dimensions, one chronological dimension, one frequency/band dimension and a final dimension reserved for representing the observer.}
       \label{fig:pSTEM}
\end{figure}

 The tensor space describing the electromagnetic field of perception can be described as five-dimensional tensor space, with three spatial dimensions, one frequency band (wavelength) dimension and one temporal (causal) dimension. The relationships between these tensorial dimensions are represented in figure ~\ref{fig:pSTEM}.
 
This generalized representation of sensorial/perceptual input acts as a good-enough abstraction to span several use cases, including image processing, video analytics, audio signal processing and representing telemetry/other measurements from a sensor mesh. This way, by adopting an encompassing tensorial representation of input, we have a better chance of avoiding the pollution with circumstantial assumption, model-specific biases and limiting, outdated technical conventions. 
 
 As a thought experiment, consider a hyperspectral camera of the future, which is able to capture not only the electromagnetic waves scattered by the surface of an object, but the behavior of all electromagnetic waves traversing the volume of the object, over a set interval of time (from pressing "Record" to pressing "Stop", for instance), with a given framerate. Of course, there are many reasons while such camera cannot exist for practical purposes, including but not limited to Heisenberg uncertainty principle \cite{boughn}.

\subsection*{Application to visual machine learning: Riveting the Multi-Graph in Latent Spaces}
In order to allow a smooth, continuous trade-off between exact matching (eg. the exact identity of two integers or the exact identity of all values in an array/string) and approximate, stochastic, probabilistic matching (eg. matching object/scene based on a nearest-neighbour algorithms) we propose a hybrid model, where each node can have one or several attributes.

{
Therefore, let us imagine that a multi-graph where each node has at least one  attribute which represents the latent-space embedding of an input image (represented as a multi-dimensional tensor, such as the pSTEM example provided above). The latent-space embedding may be obtained through whatever computation means, digital, analog or quantum, using some deep neural network architecture or a similar dimensional-reduction estimator and can be assume to be a vector of given, fixed dimension or, more generally, a tensor with a given dimensional shape and local topology.
}
{
In this multi-graph, the nodes represent a means of disposing perceived, output, processed or acquired entropy (i.e. the raw data) in some topology that ensures the desired quality attributes, such a timely availability of certain type of lookup queries, efficiency of frequent traversals or transversals of the multi-graph, at least alongside specific sequences of node type and edge type.
}
{
In the figure ~\ref{fig:typesOfEdgesAndVerties}, you can notice a minimalistic non-trivial example of such multi-graph.
}
\begin{figure}
    \makebox[\linewidth]{
        \includegraphics[width=\linewidth]{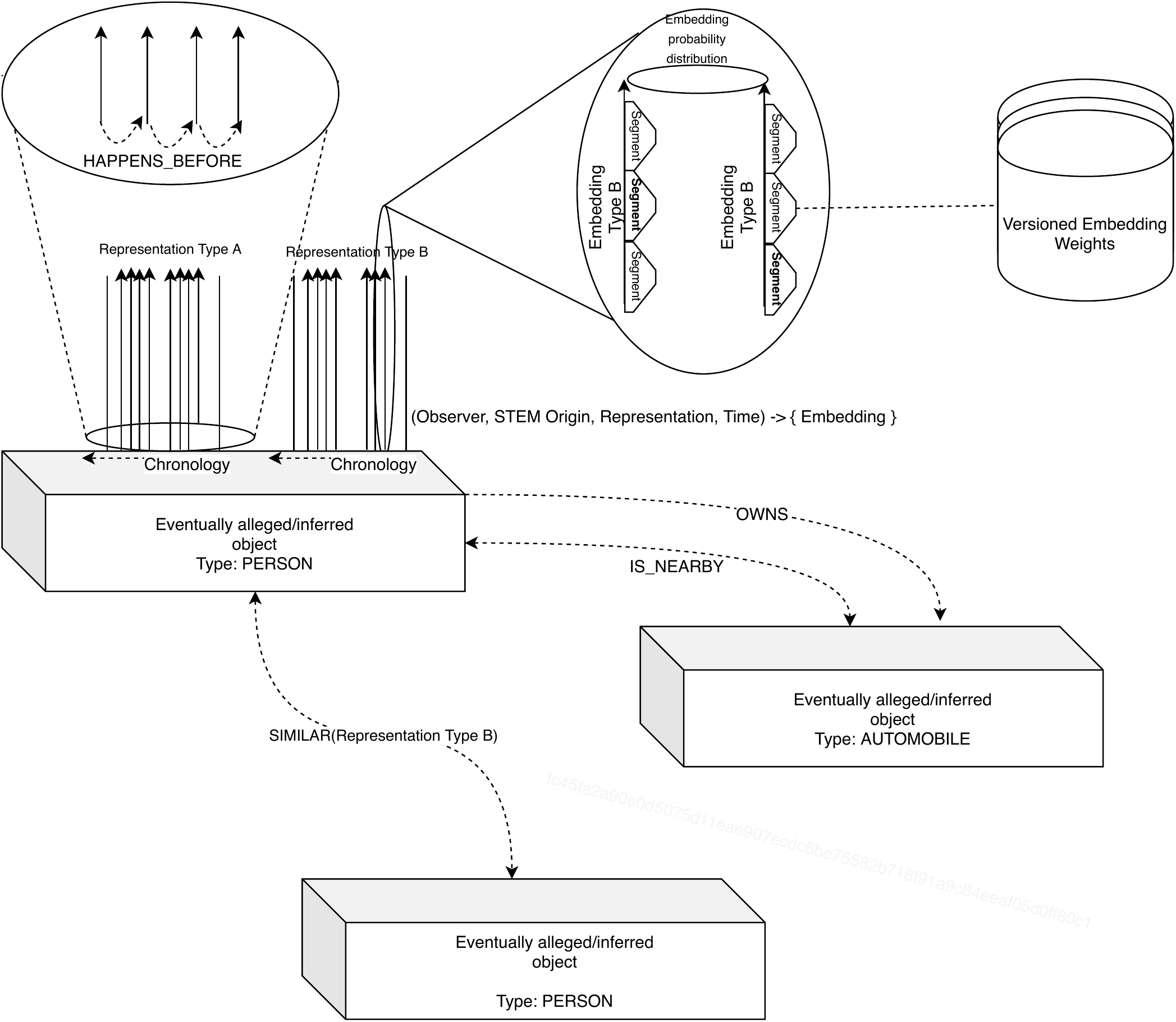}
    }
    \caption{Example utilization of multi-graph with vectorially- or tensorially-typed attributes for representing the result of video analytics, face recognition and surveillance: structures like the ones are used for internal representation of live face recognition and surveillance solutions, especially of the type-hierarchy and inclusion-structure of. Representation multi-graph building blocks. Each node contains several sets of extracted features, ordered chronologically and grouped by type. Latent space representation may have its own internal type/attribute/latent/residual representation and Coxeter-Dynkin \cite{coxeter} structure.}
    \label{fig:exampleUseCase}
\end{figure}

The measured Bhattacharyya distance between two histograms corresponding to two samples is looked up in a *big ante/post table* describing the tolerances to be usually expected from that particular sort of sampling (instrument, instrument configuration, measurements of other instruments).

\subsection*{Common types of edges $T_e$}
To make imagining practical and technical applications of the tensor/typed multi-graph more readily-available, we have compiled a short list of edge types (i.e. types of relationships). Please note that some of these edge types are generic in either one or two executable variable functions. For example, the expression \textit{"they are probably brothesr, they have the same <Shape of face> in <black and white photos>"} is one instance of the "IS\_SIMILAR\_AS\_<METRIC>\_ON\_<FIELD>" relationship/edge-types, equipped with some metric of similarity shared by several observers and observables alike ("shape of face") and with some mention/restriction/indication of the spectral field in which the perception originated ("black and white photos"). Surely, the shape of their faces might not look similar at all given a color photo, a live video or a live video with depth projection.
 The inexhaustive list of generic edge types:
 
 \begin{itemize}
 {
 \item {\textbf{HAPPENS\_BEFORE} : chronological, performed by comparison of value of a central, accepted clock or by analysis of vector clocks \cite{lamport}}
\item {\textbf{IS\_SIMILAR\_AS\_<METRIC>\_ON\_<FIELD>} }
\item {\textbf{SPATIALLY\_CONTAINS} refers to spatial inclusion of objects}
\item {\textbf{SPATIALLY\_OVERLAPS} refers to spatial overlap, intersection or tangential}
\item {\textbf{CATEGORICALLY\_CONTAINS} refers to a membership to a type-node or a category-node}
\item {\textbf{BELONGS\_TO}, \textbf{IN}   refer to membership of an object to a set, collection or enumeration }
\item {\textbf{OWNS} refer to the acknowledge ownership of one node by another node. This relationship can be considered complementary to IN and BELONGS\_TO}
\item {\textbf{IS\_$<DIRECTION>$\_PART\_OF} generic spatial inclusion given a name direction or cross-section}
\item {\textbf{IS\_LARGER\_THAN\_BY\_$<.ATTRIBUTE>$} generic comparison edge based on a comparable attributes of each node}
\item {\textbf{IS\_SEQUENCED\_AFTER\_BY\_$<COMPARATOR()>$} generic sequentiation of nodes based on a named, recursively-enumerable function of attributes of each node}
\item {\textbf{IS\_SEQUENCED\_AFTER\_BY\_$<ORDINATOR( )>$}  generic sequentiation of nodes based on a named ordinator, which represents a strict order for applying comparators on the attributes of the nodes. Examples of such ordinators would be the Unicode Collation Algorithm (UCA), based on the Default Unicode Collation Element Table (DUCET) }
}
\end{itemize}

\subsection*{The Erdos-Kirchhoff multi-graphs. Typed flows and flowing entropic cargo.}
{
In order to redefine the much needed notion of flow, we enrich a conventional multi-graph with a function of typed-flow of a specific transportable-observable (i.e. \textit{cargo}) from a node A in the graph to a node B in the multi-graph. As a homage, we label such multi-graph equipped/endowed with a typed-flow function $Flux(A,B, cargo)$, as the \textit{Erdos-Kirchhoff} typed multi-graph}
{The Erdos-Kirchhoff multi-graph can be of two kinds:}
\begin{itemize}
\item {dense, with \textit{compact} and \textit{complete} structural topology:  all possible vertices exist between neighboring nodes. This can also be thought of as a dense tesselation or a multi-dimensional, non-orthogonal lattice.}
\item {sparse or rare, without a global topology, possibly with type constraints}
\end{itemize}

The attachment of the \textit{typed-flow function} implies the ability to enforce inbound and outbound constraints on throughput using high-level stochastic computational method, in a similar way in which the flow of many electrons is modeled by Kirchhoff's equations.
Such duality between dense regions of the space, where full-mesh connectivity is more efficiently processed by a GPU or another similar shared-memory processing-core array, connected by loose/sparse edges/connections leads the way into presenting unified models, which can endure both particle constraints and fields constraints. The fitness of such models for both natural dynamics (fluids, electromagnetic waves, gravity waves) and sentient dynamics (adaptive perception, associative memory, cognition and choice) makes them a more suitable choice of implementation in the agent of machine learning, automated intelligence and generated latent representations. Moreover, the dual nature of the system, which periodically reconciles, mediates and/or any differences between the field and the particle (node) models, thus resulting in higher multi-modal robustness for real-world applications.

\subsection*{Representing Edges and Nodes in Superposition}
\begin{figure}
    \makebox[\linewidth]{
        \includegraphics[width=\linewidth]{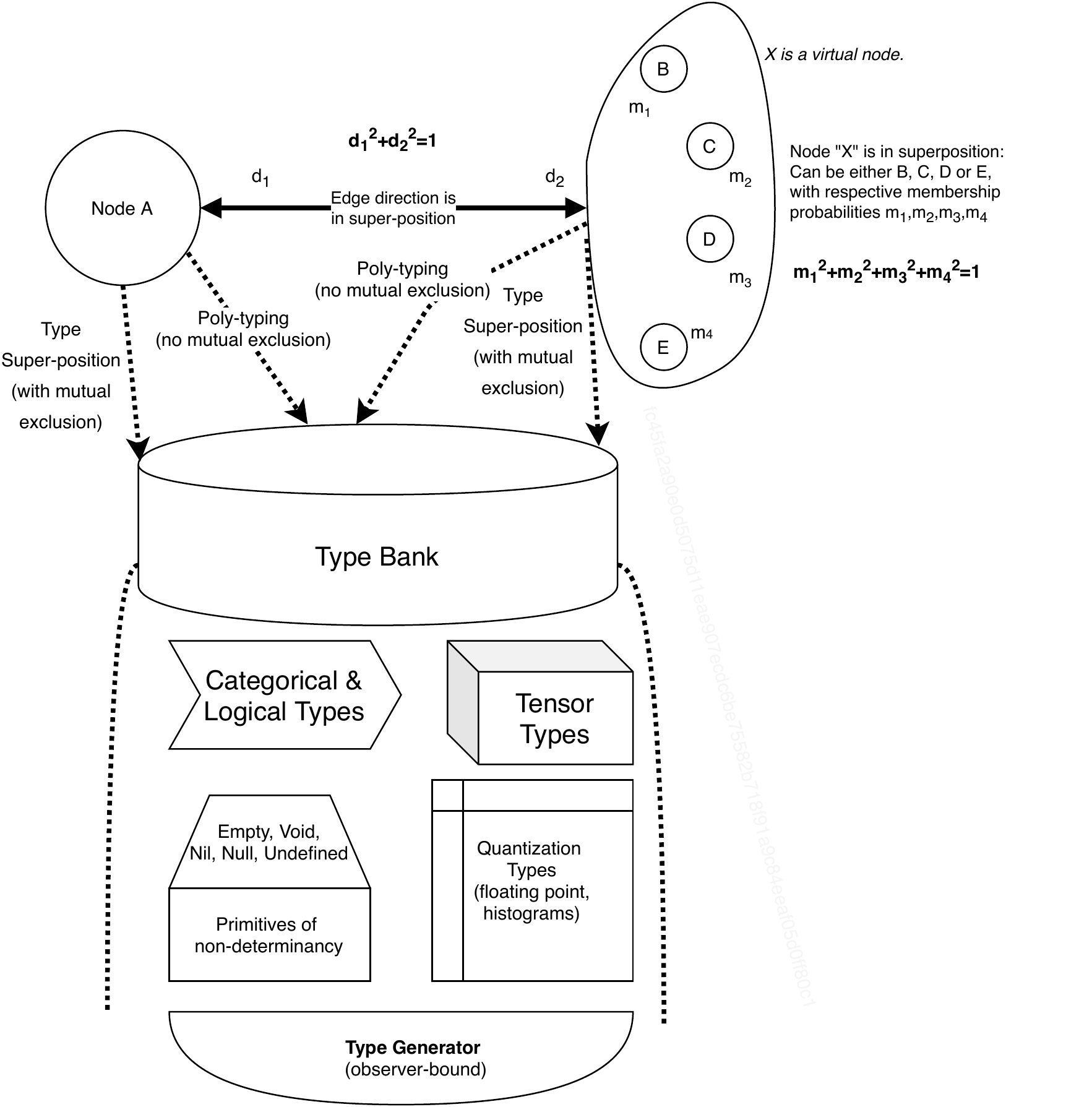}
    }
    \caption{generalizing the concept of “edge” in a graph from a deterministic object to a richer object, defined by direction, type and identity superposition.}
    \label{fig:QuantumEdge}
\end{figure}

As you can notice in Figure ~\ref{fig:QuantumEdge}, while Node A, on the lefthand side, is deterministic, the virtual node X (which is a super-position of B,C,D,E), with bounded probabilities, is non-deterministic.

The same principle of superposition can further be continuated to representing the “direction” of the relationship. While in deterministic graphs, the direction is either A->B, B->A or A<->B. In the case of an edge direction in superposition, only a normalizing quadratic constraint (similar to softmax) is placed on “directions” manifested.

\subsection*{A Generalized : The Hyper-Histogram}

The Hyper-histogram (or "hypergram", for brevity) is a heterogenous, versatile, eventually-partially orthogonal data structured which seeks to use out efficient representations of observable information in a computationally-tractable manner that is also measurable. In plainer words, the Hyper-histogram is a data structure that attempts to maximize ante-post observability (in pursuit of the principle of eventual-completeness of the graph) while minimizing the number of bits required for its representation on the Host Machine,  \textit{ as a Computationally-tractable Representation of Network of Observations}. This generalized data structure may also contain \textit{a posteriori} versioned vary-encoding (i.e. \textit{varible encoding}, also "variencoding", "vary-coding") for \textbf{typed} cells, potentially varied encoding between cells) and with cells connected either in a structured, partially-structured or unstructured fashion.

First, we need to consider the nature of the lattice cell (i.e. cellular structure of lattice), which defines a node or a vertex in the general multi-graph (multiple types of nodes and edges, only second order edges) or hyper-graph (multiple types of nodes and edge, with second-or-higher order edges):

\begin{itemize}
    \item scalars (as a degenerate case)
    \item probability distribution functions (i.e. \textit{ histograms})
    \item tensors 
    \item and other vector or abstract spaces, with at least a ring structure 
\end{itemize}

Another effect we need to concern ourselves with is the update latency and with the potentially-partial/non-atomic visibility across the changes. Cell update may be "eventually-causally consistent", in a way similar to which the \textit{AtomicLong} and \textit{AtomicInteger} are implemented in the Java specifcation \cite{java-atomic}: using several causally unmediated (independent) accumulators, which are only \textit{eventually} reconciled into the causally observable global value.

Another matter that concerns us about a computationally-traversable space is to have it endowed at least with some of the attribute that will follow, as properties of topology, measure and norm which ultimately defines the structure, the curvature and the topology of the space. One more intuitive way to think about the ramifications of the topological structure of a multi-graph is to consider the attributes listed below:

\begin{itemize}
\item \textbf{metric dimensionality} or quantization dimensionality (addressable attributes, quantized with a certain epsilon-machine function (function which makes known minimum error of the machine) )
\item \textbf{connectional dimensionality} - also defines cell connectivity (connectivity), by enumerating the neighbours with which there is at least one connection 
\item \textbf{fractal dimensionality}, which may be introduced so as to represent variable, non-local levels of connectivity between points. This is a feature of common interest in the proximities (vecinities) of fractal boundaries (frontiers), which seems to pathologically violate our assumptions about the smoothness and uniformity of the unimaginatively "smooth" spaces, such as Hausdorff spaces.
\item \textbf{curvatures} - defined by variation in angle between edges, per each pair of edge types
\item \textbf{density} - increasing or decreasing density of cells, corresponding to an decrease or increase in node size and an increase or decrease in edge length, respectively (what is the relationship between density, node size and edge length?)
\item cross-field connectivity (typed connections/edges/links between nodes/cells of *different* types) is useful when studying interaction between several fields (temperature, pressure) which have the same underlying topological-structure/support-lattice
\item { \textbf{ intra-field (\textit{intra-type}) curvature}, junction tension and differential curvature, as measured between all combinations of 2, 3, ... (up to the degree of the hyper-graph, i.e. the maximum number of nodes that can be members of the same edge) of node-type of each edge-type 
\item {link/edge type and edge category; relationship to other categories of edges. This can be useful to model statements, assertions and assumption such as \textit{"most people who <are married> have more sex between themselves"}}}
\item{ edge symmetry and assumptions of edge-traversal symmetry, so as to be able to model and to represent \textit{non-conservative systems} and intuitions such as \textit{"a two-mile road uphill is longer than a two-mile road downhill"}}
\item edge normalcity (edge normality). A normal (hyper-, multi-)graph is a graph the structure of which uniformly follows a predefined tile or tessellation. Therefore, in such graph, the topology of which would how such assumption
\item node observational spin/whirl, per each dimension or per each edge type, allows us to model and represent the order in which neighboring nodes will observe changes about others. In a simplified way, you can imagine this as local topological parsing order.
\item source of entropy for node, which represent the generalized equivalent for the cryptographic keys required for "reading" the node. This can be used for public-private key pairs, random number generators or external vaults of trust (such as a Hardware Security Module), any of them identified by a uniform resource locator, by a point or by some regular-expression or regular-expression-generator, set of materialized functions used by expression and ordered list of seeds (the seed vector, with quantized elements).
\end{itemize}
While the global tesselation and the local topology of spaces can both be defined by using Coxeter-Dynkin diagrams or Schläfli matrices (as defined in \cite{coxeter}), it is worth reviewing the intuitionistic attributes of connectivity, dimensionality, density before moving on to pursuing the formal definitions and computable models.

\subsection*{Contractions to Equality: the Bhattacharyya distance between probability distributions}
The field of machine learning is filled with examples of \textit{ dimensional reduction (DR) }, meaning a method which ingests a high-dimensional representation of an object and outputs a low-dimensional representation, which respect some constraint, usually in the form of a loss function which aims to embed a known relationship between object (eg. is same person, is same car, is same make/model of car) into the latent space representation of the sensory/perceptual tensorial volume. 
Let us consider two observable $\tilde{A}$ and $\tilde{B}$, which cannot fully or directly inspected by the observer. Instead, they can be observed through  a finite array of dimensionally-reducing (DR) maps $f_i(\tilde{x}$, each of which outputs to a space equipped with at least one distance metric $d_i(f_i(x), f_i(y))$. Let us further assume that the observer can call some generic identity (equality) function, noted   $\overset{1}{\equiv }$, to determine whether two references/instances refer to the same observable (i.e. $\tilde{A} \ \ \overset{1}{\equiv } \ \tilde{B} $). This equality function may be an external oracle, a mechanism of trial-and-error (where the Universe acts as an oracle, but with the delay of performing the experiment) or it may be be based on some assumption of spacetime continuity. As discussed in the following chapter, the intrinsic assumption over objects in the macroscopic Universe is that they don't instantaneously appear, disappear, merge, split or teleport. 

Thus, for each projector (DR map) $f_i(\tilde{x})$, we can determine a threshold of confidence $t_i$ from the equation.
\begin{equation}
  \begin{array}{l}
\int ^{t_{i}}_{0} p\left(\tilde{A} \ \ \overset{1}{\equiv } \ \tilde{B} \ |\ d_{i}\left( f_{i}\left(\tilde{A}\right) ,f_{i}\left(\tilde{B}\right)\right) < t\right) dt\ =\\
\beta +\alpha \ \int ^{\infty }_{t_{i}} p\left(\tilde{A} \ \ \overset{1}{\equiv } \ \tilde{B} \ |\ d_{i}\left( f_{i}\left(\tilde{A}\right) ,f_{i}\left(\tilde{B}\right)\right) < t\right) dt
\end{array}
    \label{eq:8}
\end{equation}
The reader familiar with machine learning and information theory will notice that, for the particular case $\alpha=1$ and $\beta=0$ we would recognize the definition for equal error rate (EER), defined as the threshold for which Type I errors are equally-frequent as Type II errors. A more in-depth analysis to how this concept can be generalized to include sensitivity and specificity can be found in \cite{ward-powers}.

For the purpose of comparing two observable states, let us consider $d_i(p,q)$ as the Bhattacharyya distance between two probability distribution, defined as:
\begin{equation*}
D_{B}(p, q)=-\ln (BC(p, q))
\end{equation*}

\begin{equation}
BC(p, q)=\sum_{x \in X} \sqrt{p(x) q(x)}
\end{equation}
Using this distance definition, we can redefine the equality operator ( "$ \equiv^{1} $" ) for essentially any type of value, including lattices with scalar or tensor elements, time series or sampled sets, as deemed relevant and with non-zero utility.

\begin{equation*}
\tilde{A} \equiv^{1} \tilde{B} <=> D_{B}(f_1(\tilde{A}, \tilde{B}) \preceq t_1
\end{equation*}

Here $t_1$ defines the threshold of the Bhattacharrya distance which is to be considered for the purpose of postulating the equality ( $ \equiv^{1}$ ) of the observable $\tilde{A}$ and $\tilde{B}$, 

We have used the "$\preceq$" to denote "less than or equal to", but also "precedes or is equal to".

We compute the $t_1$ by solving \eqref{eq:8}, with the $\alpha$ and $\beta$ deemed acceptable given the respective costs of Type I and Type II errors.

Using this constructive principle defined above, it follows that how new relationships, inferable or computable from other sources of data or with less expensive means, can be noticed, defined, validated, so as to structurally-enrich the graph not just with new data, but with \textit{ new types of relationships between existing data} , in a manner which is weakly-supervised at most and entirely non-supervised at best

\subsection*{Expansions of equality: The Dirac-Gauss operator for transforming exact equalities into convoulutions}
The notion of \textit{equality} between two entities is often overloaded, leaving confusion and room for misaligned semantics. Equality has the following implications:
\begin{itemize}
\item \textbf{ Equality of embodiment } occurs when an object $\tilde{A}$ is considered the same as an object $\tilde{B}$ because they are observed to have had \textit{corporal continuity}, which means that the motion or the path undertaken by $\tilde{A}$ to become $\tilde{B}$ has been observed to be continuous and smooth. Equality of embodiment is the principle used in  \textit{motion tracking} and \textit{optical flow}. In layman terms, this assumes that objects cannot teleport, merge into one another or divide instantenously (i.e. in the time between consecutive observations).
\item \textbf{Equality of functional potential} occurs when $\tilde{A}$ is considered the same as an object $\tilde{B}$ because they are \textit{observed to produce the same behavior} when presented with same input and context. For example, the employer-observer might observe (or assume) that employees $\tilde{B}$ and $\tilde{A}$ are equal in terms of having the same role, which means that in the context of a project requiring that role, they may be considered to be equal (at least, \textit{a priori}, before the project is started and before any performance can be measured).
\item \textbf{Equality of representation} (by an observer equipped with some observation capabilities) occurs when two objects $\tilde{A}$ and $\tilde{B}$ are considered equal based on the equality of a subset of their observable attributes. For instance, when a customs officer inspects our passport, they will observe that the facial features presented in the document correspond to the facial features of the person standing in from of them. Therefore, they will conclude, person/object $\tilde{A}$ is "the same" as $\tilde{B}$
\end{itemize}
However, in practice, there is no such thing as exact equality. The success of digital technologies is mostly reliant upon finding predictable ways to reduce representational uncertainty: making sure zeroes stay zeroes and ones stay ones, until a state change is commanded. To make the back and forth conceptual journey between exact determinism and fuzzy representation of equality/identity, let us consider the Dirac delta distribution is an edge case for a Gaussian (normal) distribution with zero standard deviation.
Thus, assuming that $\tilde{A}$ and $\tilde{B}$ are those object to be compared, we would want to compute the value of truth* for the statement $\tilde{A} ==_{[x], T_1} \tilde{B}$ 
Here $[x]$ represents the observer evaluating the statement and $T_1$ represents the definition of identity/equality used by observer $[x]$
In the general definition of the Dirac delta function 
\begin{equation}
   \int_{\mathbf{R}^{n}} A(\mathbf{z}) \delta\{d \mathbf{z}\}=\tilde{A}(\mathbf{0})
\end{equation}
 we can notice that reduces the behavior of any function that it is convolved over to the behavior of that function around the origin. This corresponds to the case where equality of objects is deterministically defined by two possible value: true and false. Note that the object $A(x)$, underlying the observable $\tilde{A}(\mathbf{0})$ is not necesarily directly and fully accessible to the observer $[x]$. Therefore, we can expand upon the semantics of the equality-identity as such, for the case of a single observer with a single observer $[x]$ :
\begin{equation}
\int_{\mathbf{U}_{[x]}} A(\mathbf{z}) \kappa_{[x], T_1}\{d \mathbf{z}\}=
\int_{\mathbf{U}_{[x]}} B(\mathbf{z}) \kappa_{[x], T_1}\{d \mathbf{z}\}
\label{eq:5}
\end{equation}

Here $\mathbf{U}_{[x]}$ represents the visible/accessible space of the observer. In a particular case of the customs officer, that would represent their visual space, which would include a representation of the passport (and its contents) and a representation of the observable person (and their face). Instead of using the Dirac $delta$ function as a kernel, we replace it with a different function $\kappa_{[x], T_1}$, which is specific to observer $[x]$ and for equality-identity of type $T_1$.

In the case of two observers who have agreed upon a joint (common) $\kappa_{ T_1}$ , we would get the following expression:
\begin{equation}
\int_{\mathbf{U}_{[x]}} A(\mathbf{z}) \kappa_{T_1}\{d \mathbf{z}\}=
\int_{\mathbf{U}_{[y]}} B(\mathbf{z}) \kappa_{T_1}\{d \mathbf{z}\}
\label{eq:6}
\end{equation}
However, in this case, making such assertion of equality between two observations performed by separate observers would furthermore require some constraint or assumption on the \textit{functional overlap} or \textit{functional equivalence} of $\mathbf{U}_{[x]}$ and $\mathbf{U}_{[y]}$
In the practical case of the customs officers, this would be equivalent to requiring that officers are human, pass an eye exam (prerequisites of the observer) and look at the subject from a prescribed distance, under prescribed lighting conditions (preconditions of the act of observation).

Using the principle of "expanding" relationships of equality, we can automate the processes by which a machine analyzing such equality/identity data from a multi-graph would be able to autonomously defined new more fine or coarse grained criteria, depending on how such an analysis is focused on efficiency, accuracy or lateral depth (i.e. avoiding bias by looking at more criteria than is strictly necessary). Concretely, should the situation require that processing is done for a lot of observable that are assumed to have little diversity, similar distributions can be "snapped" into being equal with each other, given known and acceptable rates for type I and type II errors \cite{errors}. Howevever, should the diversity and the complexity of observables be higher, statements, assertions and assumption of identity-equality can be expanded using mechanism similar to \eqref{eq:5}, \eqref{eq:6}

\subsection*{Hyper-generalizations: Spinor Graphs, Memory-bus Mapping and Path-finding, Chromatic Numbers in Higher Dimensions, Hyper-graphs}

Other concepts that help us design data and representation models which are truly useful to sentient we develop as partially autonomous include:

\begin{itemize}
\item {Spinors in Erdos-Kirchhoff Graph and the particle-observer model are useful for defining the propagation-potential of a change to a node or an edge, considering the fastest propagation time of information in the network (eg. the speed of light $c$), the smallest "traversal-length" of a graph edge (eg. Planck length $L_p$), and also considering that each node/particle/element in the network \textit{will only observes its neighbourhood in a specific order-hierarchy of iterating through rotational observation/manifestation direction per each pair of dimensions, one time-tick at a time}. The \textit{spin} is this order-hierarchy of rotating the direction for the consideration (receiving, observation) and expulsion (transmitting, manifestation) of energy is the \textbf{"spin vector"} (for isotropic behavior in several dimensions) or \textbf{"string"} (for isotropic behavior in several dimensions) of either scalars (for deterministic behaviors) or histograms (for stochastic behavior). It is observed that, depending on how spin vector are or are not alligned between neighboring nodes, the flow function may exhibit lower values (disalignment) or higher values (alignment). }
\item {The Coxeter-Dynkin diagram (Coxeter graphs, as defined and used in \cite{tumarkin}) of memory and data-buses (i.e. "magistrals") as represented in spacetime. By representing the physical manifestation of logical operations (read, write) as a problem of \textit{pathfinding in a higher-dimensional space}, we can then define the objective functions by which the Host Machine can "choose" between several functionally-equivalent paths of different computational costs (one path might be faster but block more resources and consume less energy, while another path might be slower, but more environmentally-friendly)}
\item {The mapping of two Coxter-Dynkin diagram, one representing \textit{memory layout over time} and one representing \textit{bus layout (allocation) over time}. For computational efficiency on GPU architectures, it is worth noting that every Coxeter diagram has a corresponding Schläfli matrix. Study the multi-graph correspondence to compact simplex hyperbolic groups (Lannér simplices), to paracompact simplex groups (Koszul simplices), Vinberg polytopes or other hypercompact groups have been explored but not been fully determined.}
\item {The computation of heuristics on chromatic polynomial and chromatic number $\chi(G)$ on higher dimensional tesselations and on multi-graphs. Specifically, it is of interest how does graph coloring expand to tensor-products or lexicographic-products of graphs or other multi-graphs /hyper-graphs? This is of especial  interest as Hedetniemi’s conjecture, which stated that $\chi(G \times H)=\min \{\chi(G), \chi(H)\}$, has recently been proven to be false, with a counter-example provided by \cite{shitov}. This opens the quest for lower chromatic numbers in structures obtained by joining two collection.}
\item {Generalization to multi-hyper-graphs, which involves the relationship defined by an edge no longer encloses a pair of two vertices, but a tuple involving several edge. Furthermore, here we can distinguish two cases:  hyper-graphs with tuple-edges (i.e. ordered lists), which have entropy-potential $O(2^n)$; and  hyper-graphs with set-edges (i.e. sets) complexity $O(n!)$ . This is useful for representing visual or narrative scenes with multiple subjects and/or multiple/objects, connected by the same co-occurring relationship or action. Use cases include representing relationships known or inferred between words in a sentence or representing relationships known or inferred between objects in a scene (eg. a scene from a comic strip).}

\end{itemize}

\subsection*{Availability, Durability, Replication and Multi-functional materializations}
While this work has focused primarily on the logical aspects of storing such multi-graphs, we want to bring to attention some considerations regarding the physical storage, durability, availability, partitioning and replication, as listed below:
\begin{itemize}
\item Availability (colloquially "hotness") of primitive node entropy
\item Durability is expressed as the probability that the information for inspection after $p( \delta t) t$. Durability increases with replication. A durability that exceeds the horizon of utility by to much runs the risks of producing capacity contention or costs that are too high. For this reason, the process of collection (disposal, destruction, deresolution) assures that objects past their intended durability are deallocated and the "holes" they leave behind are eventually compacted (reduced, defragmented).
\item Replication is expressed as the cardinal number of physical location where the information is stored. However, diversity of storage media, geographical location and network topology dependency are also characteristics that can increase durability, especially in the event of catastrophic or \textit{force majeure} events. Excessive replication increases cost, may increase latency for both retrieval and updates (for those operation requiring the quorum or consensus of replica) or it may decrease the likelihood of consistency (for optimistic locks and best-effort consistency schemas).

 \item Multi-functional Materialization refers to pre-computation of several versions of the primitive entropy, so the latency of the data in its most-frequently used forms (projections) is reduce. In conventional databases, this is often called a materialized view. This end-to-end  latency can be considered as latency to functionally-relevant-form, including latency of provisioning and executing any decoding tower/stack. 
 \end{itemize}

\subsection*{Discussion and Future Perspectives}
In this paper, we have explored methods of expanding the existing node and edge types in a multi-graph using the constraints of a feature extraction (i.e. dimensional reduction) function which takes in as parameter an sensory representation (such as an image or an audio clip) and which is trained/solved numerically using a convolutional estimator and a tripartite loss function (i.e. objective function). This method can be used to "stitch together" in typed-relationships (typed edge) the various latent representations output by each of the estimators, thus creating a "middle-ground" between the true/false world of logic and category theory and the smooth, probabilistic view of Bayesian behavior. This "stitching" allows for then creating hybrid ensambles algorithms, some of which are stochastic in natural (eg. deep neural nets) and other are logical/deterministic (recursive enumerable algorithms).

\subsection*{Code and Open Source Initiatives}

 DeepVISS is a non-profit initiative for standardizing inter-disciplinary data pipelines and integration for machine learning, computer vision and differential programming.
 You can find instructions on the corresponding websites:
 
 \begin{itemize}
  \item \href{https://github.com/deepviss-org}{https://github.com/deepviss-org}
  \item \href{https://deepviss.org/}{https://deepviss.org/}
\end{itemize}

 



\begin{interests}
 The authors are managing partners in several technology companies, as follows:
 \begin{itemize}
     \item Both authors (Bogdan Bocșe, Radu Jinga)  are co-founders of Knosis.AI
     \item Bogdan Bocșe is the managing partner of Envisage.AI
     \item Ioan Radu Jinga is the managing partner of Jiratech
 \end{itemize}
\end{interests}

\textcolor{white} {Signature: fc45fa2a90e0d5075d11eae907ecdc6be75582b718f91a9c84eeaf05d0ff80c1 }

\section*{Bibliography}
\bibliographystyle{zHenriquesLab-StyleBib}
\bibliography{06_Bibliography_Clean}

\begin{thebibliography}{11}
\providecommand{\natexlab}[1]{#1}
\providecommand{\url}[1]{\texttt{#1}}
\expandafter\ifx\csname urlstyle\endcsname\relax
  \providecommand{\doi}[1]{doi: #1}\else
  \providecommand{\doi}{doi: \begingroup \urlstyle{rm}\Url}\fi

\bibitem[Stan~Salvador(2005)]{salvador-chan}
Philip~Chan Stan~Salvador.
\newblock Fastdtw: Toward accurate dynamic time warping in linear time and
  space.
\newblock \emph{Florida Institute of Technology of Melbourne}, 2005.

\bibitem[Giorgio~Tomasi(2004)]{tomasi}
Claus~Andersson Giorgio~Tomasi, Frans van den~Berg.
\newblock Correlation optimized warping and dynamic time warping as
  preprocessing methods for chromatographic data.
\newblock \emph{Journal of Chemometrics}, 2004.

\bibitem[Vladimir(1966)]{levenshtein}
Levenshtein Vladimir.
\newblock Binary codes capable of correcting deletions, insertions and
  reversals.
\newblock \emph{Journal of Chemometrics}, 1966.

\bibitem[Stephen~Boughn(2017)]{boughn}
Marcel~Reginatto Stephen~Boughn.
\newblock Another look through heisenberg's microscope.
\newblock \emph{arXiv:1712.08579}, 2017.

\bibitem[Coxeter(1973)]{coxeter}
Coxeter.
\newblock \emph{Regular Polytopes}.
\newblock Dover, 1973.
\newblock ISBN 0-486-61480-8.

\bibitem[Lamport(1978)]{lamport}
Lamport.
\newblock Time, clocks, and the ordering of events in a distributed system.
\newblock \emph{Operating Systems}, 1978.

\bibitem[jav(2020)]{java-atomic}
Atomiclong implementation reference.
\newblock \emph{Java Platform Standard}, 2020.

\bibitem[Powers(2007)]{ward-powers}
David Martin~Ward Powers.
\newblock Evaluation: From precision, recall and f-factor to roc, informedness,
  markedness \& correlation.
\newblock \emph{Technical Report SIE-07-001}, 2007.

\bibitem[Martyn~Shuttleworth(2007)]{errors}
Lyndsay T~Wilson Martyn~Shuttleworth.
\newblock Experimental errors in research, 2007.

\bibitem[Tumarkin(2004)]{tumarkin}
Pavel Tumarkin.
\newblock Compact hyperbolic coxeter n-polytopes with n+3 facets.
\newblock \emph{Arxiv}, 2004.

\bibitem[Shitov(2019)]{shitov}
Yaroslav Shitov.
\newblock Counterexamples to hedetniemi's conjecture.
\newblock \emph{Arxiv}, 2019.

\end{thebibliography}

    

\end{document}